\documentclass[a4paper,fleqn, anonymous]{cas-dc}

\usepackage[numbers]{natbib}

\usepackage{latexsym}
\usepackage{graphicx}
\usepackage{amsmath}
\usepackage{hyperref}
\usepackage{booktabs}
\usepackage{caption}

\usepackage[dvipsnames]{xcolor}
\usepackage{subcaption}
\usepackage{arydshln}
\usepackage{todonotes}
\usepackage{float}
\usepackage{microtype}
\usepackage{booktabs}
\usepackage{caption} 
\usepackage{colortbl}
\usepackage{xcolor}

\begin{document}
\let\WriteBookmarks\relax

\shorttitle{EvoGens for Scientific Idea Generation}
\shortauthors{Anonymous}
\shortauthors{Li et~al.}

\title [mode = title]{EvoGens: A Population-Based Heuristic Search Framework for Scientific Idea Generation}                      
\shortauthors{Li et al.}

\author[1]{Xu Li}[orcid=0000-0002-5725-2677]
\ead{xul@swpu.edu.cn}

\author[1]{Hanzhe Tu}[orcid=0009-0006-8476-4826]
\ead{hanzhe0104@qq.com}

\author[1]{Xinyi Li}[orcid=0009-0008-3346-692X]
\ead{xinyi0733@outlook.com}

\author[1]{Kuncheng Zhao}[orcid=0009-0000-7448-8043]
\ead{zkcmster@163.com}

\author[2]{Xun Han}[]
\cormark[1]
\ead{hldwxhx@163.com}

\author[1]{Zhonghui Liu}[orcid=0009-0003-6555-9091]
\cormark[1]
\ead{lz_hui@126.com}

\cortext[cor1]{Corresponding author}

\affiliation[1]{
  organization={Southwest Petroleum University},
  city={Chengdu},
  country={China}
}

\affiliation[2]{
  organization={Sichuan Police College},
  city={Luzhou},
  country={China}
}

\begin{abstract}
Generating novel research ideas is fundamental to scientific progress. While Large Language Models (LLMs) show promise in assisting this process, existing approaches often exhibit semantic convergence, resulting in limited diversity and novelty. To address this, we introduce \textsc{EvoGens}, an evolution-inspired framework that recasts scientific idea generation as an evolutionary search over a population of ideas.
\textsc{EvoGens} iteratively applies rank-based mutation with differentiated retrieval planning to incorporate external knowledge, and semantic-aware crossover to fuse complementary concepts for conceptual reorganization.
A lightweight evaluation signal guides the selection process, encouraging sustained exploration while mitigating premature convergence. Extensive experiments demonstrate that \textsc{EvoGens} substantially enhances exploration capabilities compared to state-of-the-art baselines. Specifically, it improves the Novelty from 0.1 to 0.4 and the Diversity from 0.24 to 0.55, while maintaining comparable idea quality under the current automatic evaluation protocol.
These findings suggest that evolutionary mechanisms can serve as a useful framework for exploration-oriented research ideation, especially for broadening the novelty and diversity of candidate ideas under a shared automatic evaluation setting.


\end{abstract}



\maketitle

\section{Introduction}
Generating novel and meaningful research ideas is a central cognitive activity in scientific research \cite{baek_researchagent_2025}.
Researchers typically synthesize large volumes of previous work,
understand existing research directions,
identify unresolved problems,
and envision new methodological or conceptual advances
\cite{hope2023computationalinflectionscientificdiscovery,Wang2023ScientificDI}.
This process is inherently time-consuming and intellectually demanding,
and has become increasingly challenging with the rapid growth and fragmentation of scientific literature \cite{fire2018overoptimizationacademicpublishingmetrics}.
Recent advances in Large Language Models (LLMs) have demonstrated strong capabilities for knowledge synthesis by integrating effective reasoning and information retrieval techniques, suggesting promising new opportunities to leverage LLMs to support scientific idea generation in literature-intensive research settings \cite{openai2024gpt4technicalreport,ma_llm_2024,romera-paredes_mathematical_2024,trivedi2023interleavingretrievalchainofthoughtreasoning}.

LLM-assisted scientific idea generation frameworks are built upon the Retrieval-Augmented Generation (RAG) paradigm \cite{lewis2021retrievalaugmentedgenerationknowledgeintensivenlp,10.1007/978-981-95-4088-4_16}, utilizing external knowledge to contextualize generated concepts. The initial approaches mainly focused on retrieval optimization, employing structured or iterative semantic search to improve the relevance and novelty of the idea \cite{baek_researchagent_2025,si_can_2024}. Recently, the focus has shifted to complex generative strategies, ranging from planned retrieval for expanded topical coverage \cite{hu_nova_2024}, to structured schemes such as chain-based trend modeling \cite{li_chain_2024}, prompting-based reasoning \cite{wei2023chainofthoughtpromptingelicitsreasoning,meincke_prompting_2024}, and multi-agent collaboration \cite{su_many_2025,chen_enhancing_nodate}.

\begin{figure}
  \centering
  \includegraphics[width=0.9\columnwidth]{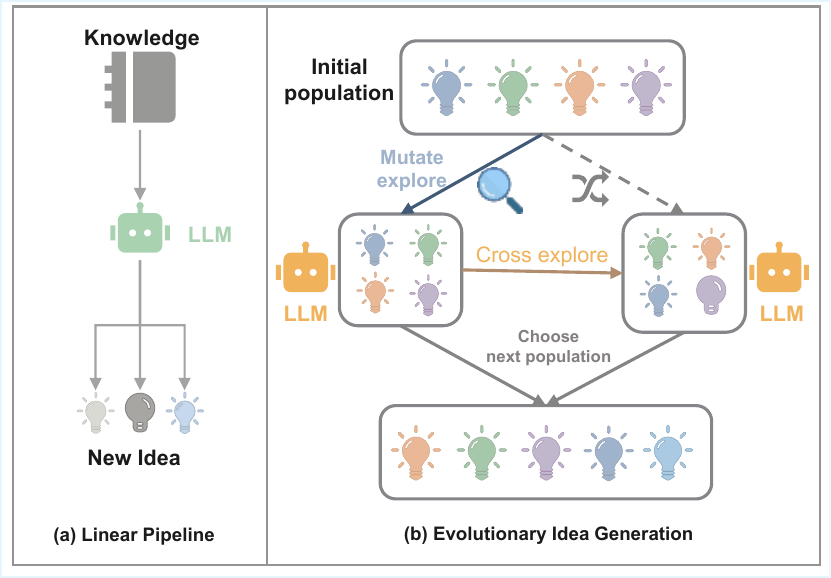}
  \caption{Paradigm comparison. (a) Conventional methods follow isolated, linear pipelines for independent generation. (b) \textsc{EvoGens} employs a population-based evolutionary process, enabling systematic exploration via iterative refinement.}
  \label{fig:evo_compare}
\end{figure}
Despite these advances, most existing methods follow a linear and isolated idea generation paradigm, as illustrated in Figure~\ref{fig:evo_compare}(a). Each idea is generated and refined independently based on its own retrieval context, with limited exploration among concurrently generated ideas.
Meanwhile, retrieval strategies in this paradigm remain predominantly relevance-driven, relying on signals such as entity co-occurrence or citation chains \cite{baek_researchagent_2025,li_chain_2024}. However, while planned retrieval strategies aim to diversify query formulation \cite{hu_nova_2024,chen_enhancing_nodate}, they are typically deployed in a static and uniform manner. This rigidity constrains idea evolution to conservative trajectories, limiting the search space. Consequently, the generation process tends to converge prematurely within existing semantic neighborhoods, weakening the diversity of research topics and the systematic exploration of cross-disciplinary research.

To address these limitations, we introduce \textsc{EvoGens}, a population-based heuristic search framework for LLM-assisted scientific idea generation. 
EvoGens initializes a diverse population of candidate research ideas and iteratively refines them via \textit{rank-based} mutation and \textit{semantic-aware} crossover.
This parallelized evolutionary process supports continuous recombination and diversification of concepts, enabling more systematic exploration of literature-grounded research directions (see Figure~~\ref{fig:evo_compare}(b)).
Taken together, \textsc{EvoGens} is intended to support early-stage conceptual exploration rather than autonomous scientific discovery.
Our contributions are summarized as follows:

\begin{itemize}
    \item We introduce \textsc{EvoGens}, a population-based framework for AI-assisted research ideation that conceptualizes early-stage idea exploration as a population-level evolutionary process. Using selection, mutation, and crossover, it transcends the limitations of linear generation, enabling systematic exploration of synergistic research directions.

    \item  We introduce the  \textit{Rank-Based Mutation (RBM)} mechanism, implemented through differentiated retrieval-plan generation, together with the  \textit{Semantic-Aware Crossover (SAC)} operator to reorganize core conceptual components.

    \item We conduct extensive experiments on a scientific idea generation benchmark. 
    Results show that EvoGens consistently improves novelty and diversity under the current automatic evaluation protocol, while maintaining comparable quality.
\end{itemize}

\section{Related Work}
\subsection{LLM-Driven Idea Generation}

Recent advances in LLMs have sparked growing interest in
AI-assisted scientific idea generation within the broader AI for Science paradigm
\cite{tang2025airesearcherautonomousscientificinnovation,luo2025llm4srsurveylargelanguage}.
Most existing work adopts retrieval-augmented generation as the core approach,
using external academic knowledge to ground the generation of new research ideas.
Representative methods construct structured retrieval mechanisms,
such as entity co-occurrence graphs or citation-based associations,
to identify relevant concepts for idea synthesis
\cite{baek_researchagent_2025,si_can_2024}.
Building upon this paradigm, subsequent studies introduce more structured retrieval
and generation strategies to enhance novelty and diversity, including
planning-based retrieval to guide query formulation and expand topical coverage
\cite{hu_nova_2024,chen_enhancing_nodate},
as well as structured or iterative generation schemes such as chain-based modeling
of research evolution \cite{li_chain_2024},
prompting-based reasoning frameworks
\cite{wei2023chainofthoughtpromptingelicitsreasoning,meincke_prompting_2024},
and multi-agent discussion settings that simulate collaborative scientific processes
\cite{su_many_2025,lu_ai_2024}.

Although these approaches have demonstrated that LLMs can generate
scientifically meaningful and sometimes highly novel ideas
\cite{Girotra2023IdeasAD,kumar2025largelanguagemodelsunlock},
they also highlight ongoing challenges in reducing redundancy
and maintaining sufficient diversity between generated ideas
\cite{si_can_2024}.
Existing iterative or retrieval-based methods
typically optimize ideas along largely independent generation trajectories, where each idea is refined in terms of its own retrieval context.
As a consequence, exploration of isolated ideas
is often implicit or indirect \cite{su_many_2025},
and systematic coordination of exploration and refinement
across multiple candidates remains underexplored.

These observations highlight the potential of population-based optimization for LLM-based idea generation. Evolutionary algorithms\cite{XIAO2009660}, as a prominent example, provide a principled framework for balancing exploration and exploitation through mechanisms such as selection, crossover, and mutation—offering a systematic paradigm that transcends isolated single-trajectory iterations.

\subsection{LLM-Based Genetic Algorithms}

Genetic Algorithms (GAs) are a class of evolutionary computation methods inspired by natural selection, extensively used in optimization, machine learning, and conceptual design~\cite{romera-paredes_mathematical_2024,kelly_interactive_2008,hongying_song_application_2024}. By simulating evolutionary operators such as selection, crossover, and mutation, GAs iteratively update a population of candidate solutions to explore the search space and progressively approach promising regions. Compared with single-point search methods, their population-based and parallel exploration mechanisms enable stronger exploration in complex and large-scale search spaces, helping avoid local optima while preserving diversity.

Driven by recent advances in LLMs, researchers are integrating evolutionary principles into LLM-based tasks. Evolutionary algorithms are considered as frameworks for structured generation and selection, rather than strict numerical optimization.
Previous studies demonstrate that LLMs can effectively simulate evolutionary operators such as mutation~\cite{lehman2022evolutionlargemodels} and crossover ~\cite{Meyerson_2024} in natural language spaces, providing a practical foundation for applying evolutionary search to structured generation tasks.
For example, EvoAgent~\cite{yuan_evoagent_2025} and EvoPrompt~\cite{guo_evoprompt_2025} leverage evolutionary algorithms to automatically generate task-specific agents or prompts, replacing manual design with population-based search and iterative refinement. 
In particular, Evolution of Heuristics (EoH)~\cite{liu2024evo} evolves natural-language heuristic ``thoughts'' together with executable code for automatic heuristic design, while LLaMEA~\cite{vanstein2025llame} iteratively generates and refines metaheuristic algorithms based on performance feedback.
Recent studies have further explored the close integration between LLMs and evolutionary processes, such as Guided Evolution (GE)~\cite{morris_llm_2024}, which uses LLM reasoning to guide evolutionary operations, and LAOS~\cite{10.1145/3712256.3726450}, which employs LLMs to adaptively control mutation and crossover based on online optimization feedback.

Evolutionary algorithms have also shown potential in scientific discovery tasks. FunSearch~\cite{romera-paredes_mathematical_2024} demonstrates that combining LLMs with evolutionary search can effectively explore discrete program spaces for combinatorial optimization. LLM-SR~\cite{shojaee_llm-sr_2025} and SGA~\cite{ma_llm_2024} further extend this paradigm to symbolic regression and physics-driven formula discovery, where evolutionary search in discrete hypothesis spaces is combined with LLM prior knowledge and, in some cases, continuous optimization via differentiable simulators. In general, these studies indicate that the integration of evolutionary principles into LLM-based generation can improve both the efficiency of the exploration and the quality of the solution.

Despite the success of existing LLM-based genetic algorithms, most approaches focus on well-structured generation targets such as formulas, code, or prompts, or executable heuristics, where candidate solutions lie in a relatively constrained search space, fitness functions are directly quantifiable, and optimization objectives are explicit. 
This is also the case for representative LLM+EA systems such as EoH and LLaMEA, which target automatic heuristic or algorithm design under explicit performance feedback.

In contrast, when generating open-ended scientific ideas in natural language,  candidate solutions have diffuse semantic boundaries, evaluation is inherently weaker, and naive crossover or mutation can introduce hallucinations, semantic redundancy~\cite{si_can_2024,hu_nova_2024}, or premature convergence.
These differences motivate evolutionary strategies tailored to literature-grounded scientific idea generation, including diversity-aware crossover for recombining semantically distinct ideas and retrieval-guided mutation for grounding modifications in relevant knowledge.

\subsection{Reasoning and Planning}
The planning and reasoning capabilities of LLMs
have become an active research focus for complex generation tasks.
Classical approaches such as Chain-of-Thought (CoT) prompting~\cite{wei_chain--thought_2023}
explicitly model intermediate reasoning steps,
while Tree-of-Thought (ToT)~\cite{yao2023treethoughtsdeliberateproblem}
extends this paradigm by enabling multi-branch exploration and self-evaluation.
For knowledge intensive settings, retrieval has been shown to further enhance reasoning effectiveness.
Representative frameworks such as Retrieval-Augmented Reasoning ~\cite{cheng2025dualragdualprocessapproachintegrate}
and ReAct~\cite{aksitov2023restmeetsreactselfimprovement}
integrate reasoning, planning, and external tool use,
allowing models to access and incorporate external knowledge during generation.
Together, these studies demonstrate that structured reasoning
and retrieval-aware planning can substantially improve generation quality
in complex tasks.

These paradigms have also been explored in idea generation and hypothesis discovery
tasks~\cite{hu_nova_2024,chen_enhancing_nodate}, where retrieval and planning are
typically used to improve individual generations under a shared strategy.
However, such uniform strategies fail to account for the heterogeneous states and
varying maturity levels of candidate ideas within a population, where different
ideas may require different degrees of knowledge grounding and semantic expansion.
This limitation suggests the need for more adaptive retrieval and planning
mechanisms in open-ended idea generation.

Taken together, prior work suggests that LLMs can support idea generation through
structured generation, retrieval, reasoning, and planning, while evolutionary
methods provide a natural mechanism for exploration and recombination.
However, existing approaches still lack a unified framework that combines
retrieval-aware planning with population-level evolutionary search for
literature-grounded research ideas.
Most current methods either follow linear or isolated generation pipelines, or
apply optimization strategies primarily designed for more structured search targets
with clearer evaluation signals.
In contrast, research ideas are open-ended natural language objects with diffuse
semantic boundaries and weakly defined fitness, requiring both sustained
diversification and adaptive control.
This gap motivates EvoGens, which integrates retrieval-aware planning with
population-based evolutionary exploration through rank-based mutation and
semantic-aware crossover, enabling continuous diversification and structured
reorganization of candidate ideas.

\begin{figure*}[htbp]
  \centering
  \includegraphics[width=0.9\textwidth]{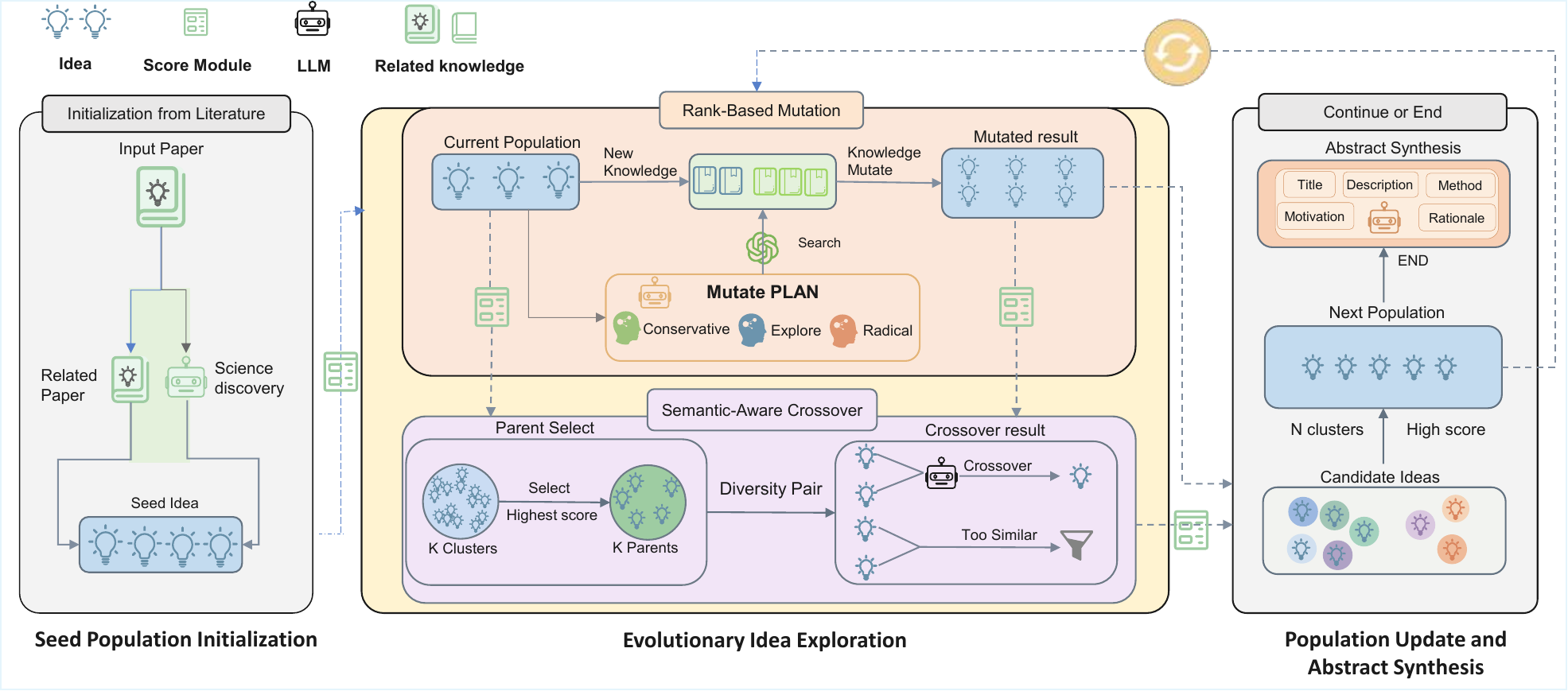}
  \caption{\textsc{EvoGens} framework.
  \textsc{EvoGens} follows a three-stage pipeline: seed idea generation,
  evolutionary idea exploration, and structured abstract generation.}
  \label{fig:evo_pipeline}
\end{figure*}

\section{METHODS}
\subsection{EvoGens Framework}

This section introduces the \textsc{EvoGens} framework, as illustrated in Figure~\ref{fig:evo_pipeline}. 
The framework consists of three evolutionary modules:
\textit{(1) Seed Population Initialization}, which constructs an initial population of candidate ideas under the research scope defined by a target paper and its reference context;
\textit{(2) Evolutionary Idea Exploration}, which updates the idea population through
the Score Module, Rank-Based Mutation (RBM), and Semantic-Aware Crossover (SAC);
and \textit{(3) Population Update and Abstract Synthesis}, which summarizes the evolved population and generates a structured research abstract for the selected ideas.

\textsc{EvoGens} formulates scientific idea generation as an evolutionary search process in natural language space. Candidate ideas are iteratively evolved through mutation, crossover, and selection.
Within this process, LLMs serve as the generative engine for seed initialization, mutation, crossover, and abstract synthesis, while a lightweight scoring model provides heuristic control for ranking, filtering, and population update. In addition, a text embedding model is used to instantiate the shared semantic space for similarity computation, clustering, and crossover pairing.
The framework is designed to support comparative search and population control within an open-ended ideation process, rather than to provide a definitive assessment of scientific validity or real-world usefulness.
Since the scorer is LLM-based, the resulting scores should be interpreted as heuristic comparative signals within a fixed evaluation setting, rather than as stable or definitive measurements of scientific merit.

\subsection{Seed Population Initialization}


The \textit{Seed Population Initialization} module constructs an initial population of candidate ideas grounded in a specific research context, providing the starting point for the subsequent evolutionary process.
Specifically, a target paper is used as the primary input, defining the research topic, direction, and domain, and thereby establishing the semantic scope for subsequent evolution.

To provide supporting context, the framework also incorporates related papers drawn from the reference context of the target paper.
These papers provide complementary background knowledge and research directions that help initialize the search space.
Based on the target paper and its related papers, the system automatically summarizes and refines relevant contextual information from the abstracts of the related papers, and uses it to guide initial idea generation.

Following Nova~\cite{hu_nova_2024}, we adopt a scientific discovery-oriented prompting strategy to encourage the generation of candidate ideas grounded in prior literature. The corresponding seed-generation prompt is provided in Tables~\ref{tab:scientific_discovery_theory} and~\ref{tab:scientific_discovery_prompt} of Appendix~\ref{sec:appendix_PROMPT}.

For the initial population, the module combines the provided contextual information with the model's internal knowledge and incorporates Chain-of-Thought reasoning to improve coherence and interpretability.
Meanwhile, self-correction and reflection mechanisms~\cite{miao2023selfcheckusingllmszeroshot} are applied to examine the internal consistency of the generated outputs, thereby reducing hallucination.
The resulting seed ideas constitute the initial population.
They are subsequently evaluated by the unified \textit{Score Module}, which is detailed in Section~3.3.
This module provides a heuristic signal for relative comparison and population control, rather than an absolute measure of scientific merit.
Based on these scores, the initial population is filtered and organized to form the first \emph{current population} that enters the evolutionary loop.

\subsection{Score Module}
Before introducing the scoring mechanism, we first define the population and structured idea representation on which the subsequent evaluation and evolutionary operators operate.

At each iteration $t$, the system maintains a population
$P_t = \{ i_1^t, i_2^t, \dots, i_N^t \}$ of $N$ candidate ideas.

Each candidate idea is represented as a structured semantic unit so that it can be evaluated and transformed at the component level during evolution.
$i = \langle T, M_{ot}, D, M_{et}, R \rangle$, corresponding to its
\emph{Title}, \emph{Motivation}, \emph{Description}, \emph{Method}, and \emph{Rationale}.
Here, \emph{Method} refers to the proposed technical mechanism or intervention,
while \emph{Rationale} captures the reasoning that explains why the proposed method
may be useful under the target problem setting.
This structured representation enables fine-grained evaluation and controlled transformation during the evolutionary process.

A unified scoring signal is needed to coordinate \textit{mutation}, \textit{crossover}, and \textit{selection} within the evolutionary process.
Rather than operating in a purely stochastic manner, these operators are coordinated
through explicit scoring signals, allowing the system to promote exploration and diversity
while maintaining basic idea quality.

At each iteration, \emph{all} ideas in the population are evaluated by a unified
\textit{Score Module}, which assigns a scalar score to each idea based on its content.
The resulting score serves as a heuristic fitness proxy for within-population control,
rather than a validated estimate of scientific merit.
It provides a unified reference for relative comparison and decision-making within the population,
including ranking, filtering, and the control of generation strategies in mutation and crossover.
Importantly, the \textit{Score Module} itself is not an evolutionary operator.
Specifically, we employ an LLM-based scorer to assess each idea along six complementary
dimensions: \emph{Innovativeness}, \emph{Feasibility}, \emph{Significance},
\emph{Clarity}, \emph{Effectiveness}, and \emph{Generalizability}
\cite{baek_researchagent_2025,li_chain_2024,qi2023largelanguagemodelszero,
su_many_2025,wang2025scipipllmbasedscientificpaper}.
The corresponding scoring prompt is provided in Table~\ref{tab:idea_eval_prompt} of Appendix~\ref{sec:appendix_PROMPT}.
These dimensions reflect the originality of the idea, the plausibility of its implementation,
its potential impact, the clarity of its presentation, its expected ability to address the target problem,
and its applicability beyond a narrow setting, respectively.
Together, they provide a coarse-grained yet informative signal for evaluating early-stage scientific ideas,
where precise quantitative metrics are often unavailable.
Each dimension is scored on an integer scale from 1 to 10, and the overall idea score is computed as a weighted linear combination (Eq.~\ref{equa1}), where $s_k(i)$ denotes the score of idea $i$ on the $k$-th dimension,
and $w_k$ is the corresponding predefined weight.

\begin{equation}
\text{Score}(i) = \sum_{k=1}^{6} w_k \cdot s_k(i),
\label{equa1}
\end{equation}

\begin{table}
\centering
\small
\setlength{\tabcolsep}{6pt}
\caption{Idea evaluation dimensions and weights.}
\label{tab:scoring-weights}

\begin{tabular*}{\columnwidth}{@{\extracolsep{\fill}}p{0.52\columnwidth}cc@{}}
\hline
\textbf{Dimension} & \textbf{Weight} & \textbf{Score Range} \\
\hline
Innovativeness  & 0.25 & 1--10 \\
Feasibility     & 0.20 & 1--10 \\
Significance    & 0.20 & 1--10 \\
Clarity         & 0.15 & 1--10 \\
Effectiveness   & 0.10 & 1--10 \\
Generalizability& 0.10 & 1--10 \\
\hline
\end{tabular*}
\end{table}

As summarized in Table~\ref{tab:scoring-weights}, the weighting scheme reflects relative priorities among different characteristics of scientific ideas,
rather than defining a universal or optimal evaluation function.
Following prior work on multi-dimensional assessment of research ideas
\cite{baek_researchagent_2025,keya2025sciideacontextawarescientificideation},
higher weights are assigned to \emph{Innovativeness} and \emph{Significance},
as \textsc{EvoGens} primarily targets the exploration of novel and potentially impactful research directions.
The remaining dimensions serve as supporting criteria that encourage basic soundness,
interpretability, and extensibility, thereby preventing incoherent or degenerate ideas
from dominating the population.


We do not claim that this weighting scheme is uniquely optimal; rather, it provides a practical heuristic configuration for within-population control in early-stage idea exploration.
The specific weight values are used as practical coefficients for introducing sufficient score differentiation among candidate ideas, which supports effective selection and controlled generation within the evolutionary loop~\cite{Zhang_Han_Ahmed-Kristensen_2025}.
We do not claim that this weighting scheme is uniquely optimal; rather, it provides a practical heuristic configuration for within-population control in early-stage idea exploration.

While the resulting scores are analyzed as continuous variables, they are discretized into coarse quality tiers
during specific evolutionary stages.
This discretization supports subsequent rank-based and crossover-related control mechanisms,
as detailed in the following sections.

\subsection{Rank-Based Mutation (RBM)}

Rank-Based Mutation (RBM) assigns different mutation modes to ideas according to their relative ranking in the current population.
Building on the unified scoring signal introduced above, we first rank ideas
within the current population and then partition the ranked list into three
coarse tiers, which are mapped to different mutation modes.
We refer to this mechanism as \textit{Rank-Based Mutation} (RBM):
the score determines relative ordering among ideas, while the actual mutation
behavior is assigned according to rank intervals rather than absolute score thresholds~\cite{wang2025largelanguagemodelsmeet}.

Operationally, RBM transforms an input idea into a literature-grounded variant through retrieval-plan generation, literature retrieval, and conditioned rewriting.
From an operator perspective, RBM is implemented through retrieval-plan generation,
external knowledge acquisition, and conditioned rewriting, transforming an existing
idea into a literature-grounded variant.

Formally, the mutation process first generates a retrieval plan, then retrieves external literature, and finally synthesizes a mutated idea conditioned on the retrieved knowledge.
Operationally, RBM takes an input idea, applies rank-conditioned retrieval planning and diversity control, and outputs a literature-grounded mutated idea.

Given a current idea $i$, the system first generates a mutation-specific retrieval plan
under a planning regime $\rho$ and diversity control $\tau$:
\begin{equation}
p \sim \mathcal{P}(i; \rho, \tau),
\label{equa2}
\end{equation}
where $\mathcal{P}(\cdot)$ (Eq.~\ref{equa2}) denotes the LLM-based retrieval-plan generation process, and the resulting
plan $p$ specifies the search fields and keywords used for external knowledge acquisition.
The system then executes this plan through an academic search engine\footnote{In this work, literature retrieval is implemented using the Semantic Scholar API.}  to obtain
retrieved literature:
\begin{equation}
k \sim \mathcal{R}(p),
\label{equa3}
\end{equation}
where $\mathcal{R}(\cdot)$ (Eq.~\ref{equa3}) denotes literature retrieval conditioned on the generated plan.
Finally, the retrieved knowledge $k$ is used to generate a mutated idea via (Eq.~\ref{equa4})
\begin{equation}
\mathcal{M}_{ut}(i, k) = \mathrm{LLM}(i, k).
\label{equa4}
\end{equation}

Unlike classical genetic algorithms, where mutation is often implemented as a low-probability
random perturbation, \textsc{EvoGens} treats mutation as a primary knowledge-acquisition
operator for exploring an open-ended idea space.
Each mutation is grounded in newly retrieved academic literature, allowing the system
to continuously introduce external knowledge and reduce semantic stagnation under limited
initial context.
Accordingly, mutation is applied to \emph{all} individuals in the current population
before crossover, so that recombination operates on ideas that have already been enriched
through knowledge-driven expansion.
Therefore, the mutation stage should be understood as retrieval-grounded, selection-guided variation rather than unconstrained random perturbation: diversity is introduced through differentiated planning and external literature acquisition, while remaining bounded by the current idea and retrieved evidence.

In implementation, the ranked population is partitioned into three coarse tiers using
a 2:2:1 ratio.
The top-ranked ideas are assigned to \textit{Conservative} mutation, the middle-ranked
ideas to \textit{Explorative} mutation, and the bottom-ranked ideas to \textit{Radical}
mutation.
This allocation is used as a practical heuristic to balance stability and diversification,
rather than a theoretically optimal ratio.
The three mutation modes correspond to progressively stronger retrieval planning regimes
and diversity settings.
As summarized in Table~\ref{tab2}, Conservative mutation emphasizes local refinement
using closely related literature; Explorative mutation expands toward adjacent or
less-explored topics; and Radical mutation encourages broader semantic deviation by
probing semantically distant or unconventional research directions.

Notably, prompt variation and temperature control are applied primarily at the stage of
retrieval-plan generation rather than unrestricted final idea synthesis.
This design localizes the risk of higher-diversity generation:
even when a more exploratory plan produces noisy or weak leads, the resulting effect is
typically limited to failed or less relevant retrieval, rather than directly injecting
hallucinated content into the final mutated idea.
The final mutation step remains conditioned on retrieved academic literature, preserving
evidence grounding and traceability.

\begin{table}[htbp]
\centering
\small
\renewcommand{\arraystretch}{1.15}
\setlength{\tabcolsep}{3.5pt}
\caption{Overview of rank-based mutation modes in \textsc{EvoGens}. Conservative, Explorative, and Radical modes are instantiated with progressively increasing temperature settings in implementation.}
\begin{tabular}{p{0.20\columnwidth} p{0.16\columnwidth} p{0.34\columnwidth} p{0.20\columnwidth}}
\toprule
\textbf{Mutation Mode} & \textbf{Tier} & \textbf{Retrieval Scope} & \textbf{Intended Behavior} \\
\midrule
Conservative & Top-ranked & Closely related literature & Local refinement \\
Explorative & Mid-ranked & Adjacent or less-explored topics & Balanced expansion \\
Radical & Bottom-ranked & Semantically distant or unconventional directions & Broad semantic deviation \\
\bottomrule
\end{tabular}

\label{tab2}
\end{table}

At the implementation level, the three mutation modes in Table~\ref{tab2} are instantiated
with distinct retrieval-plan prompts and corresponding temperature settings, producing
low-, medium-, and high-diversity retrieval plans, respectively.
The corresponding retrieval-plan prompts for Conservative, Explorative, and Radical mutation are provided in Tables~\ref{tab:conservative_mutation_prompt}, \ref{tab:exploratory_prompt}, and \ref{tab:radical_prompt} of Appendix~\ref{sec:appendix_PROMPT}.
In our implementation, these three modes use temperatures of 0.6, 0.8, and 1.0.

For each idea, the LLM first formulates a mutation-specific retrieval plan specifying
search fields and keywords.
This plan is then executed through an external academic search engine
(in our implementation, Semantic Scholar) to obtain idea-specific new knowledge from the literature.
Conditioned on the retrieved content, the LLM generates a refined or transformed idea
using a shared mutation-generation prompt, ensuring that mutations remain evidence-based and traceable.
The corresponding mutation-generation prompt is provided in Table~\ref{tab:idea_mutate_prompt} of Appendix~\ref{sec:appendix_PROMPT}.

\subsection{Semantic-Aware Crossover (SAC)}

In this stage, \textsc{EvoGens} recombines ideas across semantically distinct
regions of the population, where semantic distinctions are determined in the
embedding space described below.
This design promotes knowledge integration while reducing redundant recombination
between highly similar ideas.
The crossover process operates on a unified candidate pool that combines
the mutated results with the current population,
allowing newly generated ideas to interact with existing ones.

To construct diverse and high-quality crossover parents, SAC first embeds the candidate pool into a shared semantic space and performs clustering-based parent selection.
Let $\mathcal{C}_t$ denote the raw candidate set at iteration $t$,
which comprises the current population and their mutated variants.
All ideas in $\mathcal{C}_t$ are embedded into a shared semantic space.
Here, the shared semantic space is instantiated as the sentence-embedding space
produced by all-MiniLM-L6-v2, and semantic distance is measured by cosine distance
between idea embeddings.
This embedding space is used only for clustering and parent pairing, rather than
as a direct substitute for the idea representation itself.
To encourage semantic diversity while retaining strong candidates in parent selection,
we apply $k$-means clustering to partition the embeddings into $k$ clusters.
From each non-empty cluster, the idea with the highest score
(as evaluated by the Score Module) is selected as a representative.
This clustering-based selection yields the parent set
$\mathcal{P}_t = \{ i_1, i_2, \dots, i_m \}$, where $m \leq k$,
ensuring that the selected parents span diverse semantic regions
while prioritizing high-quality candidates within each region.

After representative parents are selected, SAC forms crossover pairs by greedily matching semantically distant ideas.
Based on the selected parent set $\mathcal{P}_t$, multiple parent pairs are
constructed for crossover in a greedy and distance-aware manner.
For each idea $i \in \mathcal{P}_t$ that remains unpaired, we compute its semantic
distance to all other unpaired ideas and select the one with the largest distance
as its crossover partner.
Once a pair is formed, both ideas are removed from the candidate list so that each
idea participates in at most one crossover.
This procedure yields a set of parent pairs:

\[
\mathcal{M}_t^{\text{match}} =
\{(i_a^{(1)}, i_b^{(1)}), (i_a^{(2)}, i_b^{(2)}), \dots, (i_a^{(r)}, i_b^{(r)})\},
\]
which collectively support the generation of multiple offspring
within a single evolutionary iteration.

Operationally, SAC takes a pair of semantically distant parent ideas, applies a structured crossover prompt over their complementary components, and outputs a synthesized offspring idea.
For each parent pair $(i_a, i_b) \in \mathcal{M}_t^{\text{match}}$,
the crossover operator integrates their complementary conceptual components
through an LLM-based generation process:
\begin{equation}
\mathcal{C}_{\text{cross}}(i_a, i_b) = \text{LLM}(i_a, i_b).
\label{equa5}
\end{equation}
For each parent pair $(i_a, i_b)$ (Eq.~\ref{equa5}), the LLM is prompted with the structured components of both ideas (e.g., motivation, description, method, and rationale) and instructed to synthesize an offspring that recombines complementary elements into a coherent new idea.
The corresponding crossover-generation prompt is provided in Table~\ref{tab:crossover_prompt} of Appendix~\ref{sec:appendix_PROMPT}.

The generated crossover offspring are merged with the mutated ideas to form the candidate pool for subsequent population update.
Combined with clustering-based parent selection, this crossover design is intended to reduce redundant recombination while encouraging  knowledge integration.


\subsection{Population Update and Abstract Synthesis}
This module governs population-level selection and generational transition,
consolidating evolved ideas across iterations and producing the final output.

After mutation and crossover, the newly generated ideas are merged into a candidate pool from which the next population is selected.
Only ideas generated in the current iteration are eligible for selection into the next population, while ideas from the previous population are not directly carried over.
The population update step formalizes how the next generation is selected from the mutated and crossover-produced candidates at iteration $t$.

\begin{equation}
P_{t+1} = \mathcal{S}_{\text{pop}}\bigl(P_t^{\text{mut}} \cup P_t^{\text{cross}}\bigr)
\label{equa6}
\end{equation}

Here, $\mathcal{S}_{\text{pop}}(\cdot)$ (Eq.~\ref{equa6}) denotes the population selection operator that determines the composition of the next generation.

This generational replacement strategy avoids repeatedly selecting unchanged ideas
and promotes sustained semantic exploration across iterations.
To implement $\mathcal{S}_{\text{pop}}(\cdot)$ while preserving diversity,
the candidate pool is partitioned into $n$ clusters in the semantic embedding space,
and the highest-scoring idea from each cluster is selected to form the next population.
If the maximum number of evolutionary iterations has not been reached,
$P_{t+1}$ serves as the new current population for the next iteration;
otherwise, it is treated as the final evolved population.

After the final iteration, the final population of candidate research ideas
is converted into a structured abstract representation.
Following VIRSCI~\cite{su_many_2025},
an LLM maps each idea into a standardized template consisting of
\textit{title}, \textit{research motivation}, \textit{method},
and \textit{idea description},
ensuring a consistent semantic structure for downstream analysis.

\section{EXPERIMENTS}

\subsection{Data}
We evaluate our method using a dataset derived from \textit{MAGenIdeas}~\cite{chen_enhancing_nodate}, comprising 144 long papers published in ACL 2024. Each instance in \textit{MAGenIdeas} is centered on a target paper, which contains its title, abstract, and a set of referenced papers that provide an essential background context. 
This structure mimics authentic scientific discovery, where researchers formulate new ideas by synthesizing existing work with existing literature.

To ensure a fair comparison and accommodate LLMs context window constraints, we standardize the input across all methods. Specifically, each system is provided with the target paper and its top-10 semantically relevant references. This constraint normalizes the ``information budget'' available to each model, ensuring that performance differences reflect generative capability rather than data access.

\subsection{Experimental Configuration}
Within \textsc{EvoGens}, different models are assigned to generation, scoring, and semantic representation roles.
Specifically, we use GPT-4o-mini~\cite{Hurst2024GPT4oSC} as the generation model for seed idea initialization, rank-based mutation, semantic-aware crossover, and abstract synthesis.
GPT-5-mini~\cite{singh2025openaigpt5card} is adopted as the scoring model in the unified \textit{Score Module}, where it provides relative score signals for ranking, filtering, and mutation control during the evolutionary process.
Consistent with the method design, this scoring model instantiates the unified scorer used during evolution.

In addition, we use \textit{all-MiniLM-L6-v2}~\cite{wang2020minilmdeepselfattentiondistillation} as the text embedding backbone.
Its embeddings instantiate the shared semantic space used for semantic similarity computation, clustering, and parent pairing in the crossover stage.

\subsection{Baselines}
To evaluate effectiveness, we compare \textsc{EvoGens} with four representative baseline approaches: AI-Researcher~\cite{si_can_2024}, AI-Scientist~\cite{lu_ai_2024}, Future-Idea-Generation~\cite{kumar2025largelanguagemodelsunlock}, and  MAGenIdeas~\cite{chen_enhancing_nodate}.

For the first three baselines, AI-Researcher, AI-Scientist, and Future-Idea-Generation, we directly use their official open-source implementations without modifying the core generation pipelines.
AI-Researcher is evaluated using its end-to-end idea generation framework, while for AI-Scientist, we extract and evaluate only its research idea generation module, excluding other downstream components such as experiment design and paper writing.

For the MAGenIdeas, we adopt its default experimental settings, using three iterative discussion rounds with four agents. To ensure a comparable number of generated ideas across methods, \textsc{EvoGens} is configured to run for three evolutionary iterations.

\subsection{Metrics}
Following the evaluation protocols of AI-Researcher and Nova~\cite{si_can_2024,hu_nova_2024}, we examine generated research ideas from three perspectives: \textit{Novelty}, \textit{Diversity}, and \textit{Quality}.
Among them, novelty and diversity are used as exploration-oriented automatic proxies that characterize semantic expansion and population spread, rather than fully establishing the scientific usefulness of the generated ideas.

\paragraph{Novelty.}~
We adopt a semantic similarity--based approach to measure the novelty of generated ideas.
Specifically, each generated idea and its corresponding reference papers are mapped to vector representations using the same text embedding model, and the cosine similarity between them is computed.

An idea is considered novel if its maximum similarity to all reference papers is below a predefined threshold $\theta_n$.
Following previous studies~\cite{hu_nova_2024,kumar2025largelanguagemodelsunlock}, we set $\theta_n = 0.4$.
The novelty score is defined as
\begin{equation}
\mathrm{Novelty} = \frac{1}{n} \sum_{i=1}^{n} \mathbb{I}\!\left(\max_j \mathrm{sim}(a_i, r_{ij}) < \theta_n \right),
\label{n_novelty}
\end{equation}
where $n$ denotes (Eq.~\ref{n_novelty}) the number of generated ideas, $a_i$ denotes the $i$-th generated idea, $r_{ij}$ denotes its related reference papers, $\mathrm{sim}(\cdot,\cdot)$ denotes cosine similarity, and $\mathbb{I}(\cdot)$ denotes an indicator.

\paragraph{Diversity.}~
Consistent with previous works~\cite{si_can_2024,hu_nova_2024}, diversity is measured by the proportion of semantically unique ideas among the generated results.
Using the same embedding model and cosine similarity computation as in the novelty metric, we compare each generated idea against the remaining generated ideas.
An idea is counted as diverse if its maximum similarity to all other generated ideas is below a duplication threshold $\theta_d$ (Eq.~\ref{m_d}).
Following prior work, we set $\theta_d = 0.8$. The diversity score is defined as
\begin{equation}
\mathrm{Diversity} = \frac{1}{n} \sum_{i=1}^{n} \mathbb{I}\!\left(\max_{j \neq i} \mathrm{sim}(a_i, a_j) < \theta_d \right).
\label{m_d}
\end{equation}
This metric reflects population-level semantic spread rather than direct scientific quality.

\paragraph{Relative Quality.}~
The quality of generated research ideas is evaluated using a Swiss-system tournament combined with a zero-shot LLM-based ranker, following prior work~\cite{si_can_2024,hu_nova_2024}.\footnote{In our implementation, the zero-shot ranker is instantiated with DeepSeek-V3.}
Under this protocol, ideas are compared pairwise, and the ranker determines which idea is superior in each comparison.
Each idea participates in five rounds of pairwise comparison and receives one point for each win.
Let $s_i$ denote the total tournament score of the $i$-th idea.
Following prior settings, an idea is regarded as high-quality if $s_i \geq 5$.
The reported quality score in Table~\ref{tab:main_results} is the proportion of high-quality ideas:
\begin{equation}
\mathrm{Relative\ Quality} = \frac{1}{n} \sum_{i=1}^{n} \mathbb{I}(s_i \geq 5),
\label{r_qua}
\end{equation}
where $n$ (Eq.~\ref{r_qua}) denotes the total number of generated ideas.
This metric is a comparative automatic proxy derived from pairwise ranking, rather than an absolute measure of scientific merit.
Accordingly, the quality results reported in this work should be interpreted as relative evidence under a shared automatic ranking setup, not as definitive judgments of scientific validity, feasibility, or long-term impact.

\subsection{Implementation}
Considering the cost and evaluation criteria,
all methods generate 40 research ideas for comparison in this study.
\textsc{EvoGens} is executed for 3 evolutionary iterations
with a maximum population size of 40.
During parent selection,
30 parent ideas are selected by clustering k-means (k=30) to encourage diversity,
retaining the highest-scoring idea from each group.
The same clustering-based mechanism is applied during next-generation selection.
When the number of candidates is smaller than the cluster count,
all candidates are preserved.
For each evolutionary operation,
both crossover and mutation generate \emph{two} new offspring ideas.
Overall, at most 110 new ideas are generated per iteration.

\subsection{Experimental Results}

Table~\ref{tab:main_results} shows that the evolution-inspired generation framework substantially improves the novelty and diversity of the generated research ideas under the current automatic evaluation protocol.
In particular, \textsc{EvoGens} exhibits a clear advantage in novelty, which is an important criterion for assessing the potential originality of research ideas.

Compared with the baselines, \textsc{EvoGens} achieves the strongest performance on both \textit{Novelty} and \textit{Diversity}, while its quality remains competitive with strong baselines, though not superior to the best-performing quality baseline.
This result suggests a trade-off consistent with the exploration-oriented design of \textsc{EvoGens}: the framework primarily promotes conceptual exploration and diversification while maintaining a competitive level of automatic comparative quality.

\begin{table}[htbp]
\centering
\small
\renewcommand{\arraystretch}{1.2}
\caption{Overall performance comparison under the automatic evaluation protocol.}
\resizebox{\columnwidth}{!}{%
\begin{tabular}{p{0.45\columnwidth}ccc}
\toprule
Method & Novelty $\uparrow$ & Diversity $\uparrow$ & Relative Quality $=$ \\
\midrule
AI-Scientist (2024) & 0.10 & 0.24 & 0.13 \\
Future-Idea-Generation (2024) & 0.02 & 0.42 & 0.14 \\
MAGenIdeas (2025) & 0.03 & 0.17 & 0.21 \\
AI-Researcher (2025) & 0.05 & 0.38 & \textbf{0.24} \\
\midrule
\rowcolor[HTML]{CEEAF7}
\textbf{EvoGens (Ours)} & \textbf{0.40} & \textbf{0.55} & 0.21 \\
\bottomrule
\end{tabular}%
}
\label{tab:main_results}
\end{table}

\begin{table}[htbp]
\centering
\small
\renewcommand{\arraystretch}{1.15}
\caption{Turn-level evolution of novelty and diversity for the full \textsc{EvoGens} model across generations.}
\label{tab:turn_analysis}

\begin{tabular*}{\columnwidth}{@{\extracolsep{\fill}}lccc@{}}
\toprule
\textbf{Metric} & \textbf{Turn 1} & \textbf{Turn 2} & \textbf{Turn 3} \\
\midrule
Novelty   & 0.42 & 0.38 & 0.40 \\
Diversity & 0.35 & 0.53 & 0.55 \\
\bottomrule
\end{tabular*}
\end{table}

\begin{figure}
  \centering
  \scalebox{0.99}{
    \includegraphics[width=\columnwidth]{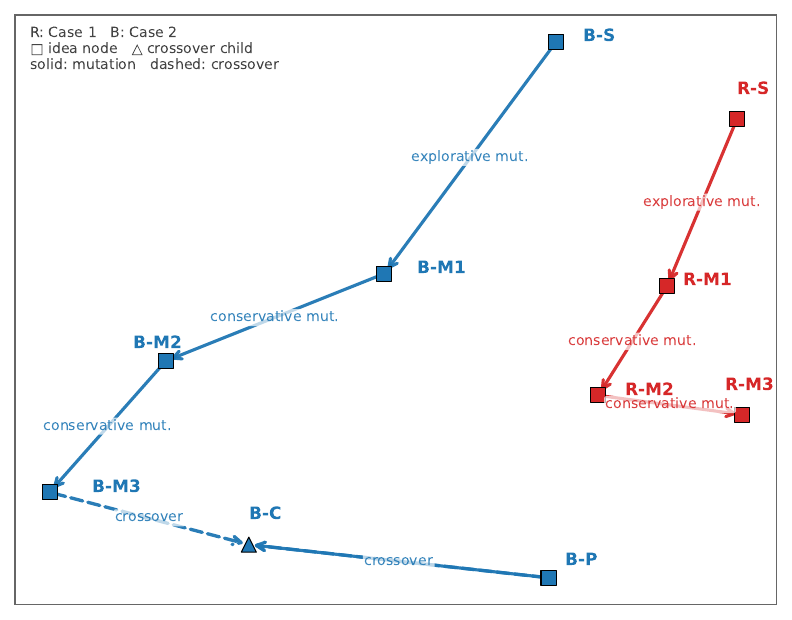}
  }
  \caption{t-SNE visualization of two representative idea evolution chains in \textsc{EvoGens}. Square nodes denote idea states, the triangular node denotes a crossover child, solid arrows indicate mutation, and dashed arrows indicate crossover. The red chain is mutation-only, while the blue chain includes both mutation and crossover.}
  \label{fig:t-sne}
\end{figure}
A turn-level analysis is reported in Table~\ref{tab:turn_analysis}, showing the evolution of novelty and diversity across generations for the full \textsc{EvoGens} model.
\textit{Diversity} increases consistently with each turn, with the largest improvement observed from Turn~1 to Turn~2 (0.35 to 0.53), while the gain from Turn~2 to Turn~3 becomes smaller (0.53 to 0.55), indicating diminishing returns in later iterations under the current budget setting.
In contrast, \textit{novelty} exhibits a fluctuating trend and shows a slight decrease in later generations, suggesting that iterative evolution gradually shifts from introducing new variations to reorganizing existing concepts through recombination.
Similar behaviors have been reported in prior idea generation frameworks such as
NOVA and MAGenIdeas~\cite{hu_nova_2024,chen_enhancing_nodate}.
Given that \textsc{EvoGens} primarily targets diversity and conceptual exploration, we adopt three evolutionary turns as a practical trade-off between exploration gain and computational cost.
At this point, \textit{diversity} has improved substantially, while novelty remains at a competitive level, providing a practical operating point for the generation of early-stage research ideas.
The weaker gains after Turn~3 do not necessarily indicate that the search space is exhausted; instead, they more likely reflect diminishing marginal returns under the current operator design and computational budget.

To qualitatively examine how ideas evolve in semantic space, we visualize two representative idea evolution chains using t-SNE, as shown in Figure~\ref{fig:t-sne}. The plot is not intended to interpret the t-SNE axes themselves, but to illustrate relative semantic displacement during mutation and crossover.

\textit{The red chain }shows a mutation-only case. It begins with an explorative mutation, which induces a relatively larger semantic shift, and is followed by two conservative mutations that produce smaller local refinements. This pattern is consistent with the intended behavior of rank-based mutation: broader exploration first, then local refinement.

\textit{The blue chain} shows a mutation-and-crossover case. A radical mutation first produces a larger semantic displacement, after which explorative mutation further reshapes the idea. The final crossover child remains semantically related to its parent ideas while occupying a distinct position, suggesting that crossover in \textsc{EvoGens} performs structured conceptual recombination rather than simple text fusion.

Overall, the two chains show that different mutation modes induce different degrees of semantic change, while crossover yields offspring that are related to, but still distinct from, their parent ideas.

\section{ANALYSIS AND DISCUSSION}

\subsection{Ablation Study}

To analyze the contribution of each component in \textsc{EvoGens}, we conduct a series of ablation studies by selectively removing the \textit{Rank-Based Mutation} (RBM) mechanism and/or the \textit{Semantic-Aware Crossover} (SAC) mechanism.
All ablation variants keep the same mutation retrieval budget ($n=6$ papers retrieved per mutation step) to ensure a fair comparison.
Specifically, \textsc{EvoGens} w/o RBM retains the retrieval-plan-based mutation process but removes rank-based differentiation, using a single shared retrieval-plan prompt and a unified temperature setting for all ideas;

\textsc{EvoGens} w/o Crossover removes inter-idea recombination entirely; and \textsc{EvoGens} w/o SAC retains crossover but replaces semantic-aware pairing with random pairing.
The results are summarized in Table~\ref{tab:ablation}, which reports novelty, diversity, and automatic comparative quality.

\paragraph{\textsc{EvoGens} w/o RBM.}~
This variant retains the retrieval-plan-based mutation mechanism but removes the rank-based mutation control strategy.
All ideas share the same retrieval-plan prompt (Table~\ref{tab:search_keyword_prompt} in Appendix~\ref{sec:appendix_PROMPT}) and a unified temperature setting ($T=0.7$).
Novelty drops dramatically (from 0.40 to 0.06), accompanied by a slight decrease in diversity (from 0.55 to 0.51), while relative quality decreases only slightly (from 0.21 to 0.20).
This indicates that the main contribution of RBM does not come from any single radical mutation mode in isolation, but from differentiated mutation control across the population.
By assigning conservative, explorative, and radical mutation modes to different rank tiers, RBM provides complementary semantic perspectives that jointly expand the search space.
Without such rank-based differentiation, mutation tends to remain more homogeneous, and crossover mainly reinforces convergence toward retrieved references rather than promoting broader exploration.


\begin{table}[htbp]
\centering
\small
\renewcommand{\arraystretch}{1.2}
\caption{Ablation study of \textsc{EvoGens} framework. RBM denotes Rank-Based Mutation, and SAC denotes Semantic-Aware Crossover.}
\resizebox{\columnwidth}{!}{%
\begin{tabular}{p{0.45\columnwidth}ccc}
\toprule
Method & Novelty $\uparrow$ & Diversity $\uparrow$ & Relative Quality $\uparrow$ \\
\midrule
w/o RBM & 0.06 & 0.51 & 0.20 \\
w/o RBM \& Crossover & 0.05 & 0.49 & 0.19 \\
w/o Crossover & 0.25 & 0.51 & 0.16 \\
w/o SAC & 0.40 & 0.52 & 0.17 \\
\midrule
\textbf{Full \textsc{EvoGens}} & \textbf{0.40} & \textbf{0.55} & \textbf{0.21} \\
\bottomrule
\end{tabular}%
}
\label{tab:ablation}
\end{table}

\paragraph{\textsc{EvoGens} w/o RBM \& Crossover.}~
This variant removes both rank-based mutation control and crossover while retaining a uniform retrieval-plan-based mutation process.
All ideas share the same retrieval-plan prompt and temperature setting, with no inter-idea recombination.
Both novelty and diversity decline substantially (from 0.40 to 0.05 and from 0.55 to 0.49, respectively), while relative quality also decreases from 0.21 to 0.19.
Compared with \textsc{EvoGens} w/o RBM, removing crossover does not reduce novelty further, but it does reduce diversity and relative quality.
This suggests that without RBM, crossover contributes little to exploration, yet still helps preserve the overall coherence and competitiveness of generated ideas under the current automatic evaluation protocol.


\paragraph{\textsc{EvoGens} w/o Crossover.}~
This variant completely removes the crossover operator while keeping the RBM process intact.
All ideas evolve solely through rank-based mutation without inter-idea recombination.
As a result, both novelty and diversity decrease (from 0.40 to 0.25 and from 0.55 to 0.51, respectively), and relative quality also drops from 0.21 to 0.16.
This indicates that differentiated retrieval planning alone is sufficient to introduce substantial semantic divergence even without explicit idea recombination.
At the same time, the decline in relative quality suggests that crossover helps preserve coherence and integration across generated ideas, rather than simply adding novelty.

\paragraph{\textsc{EvoGens} w/o SAC.}~
This variant retains the crossover operator but removes the semantic-aware pairing strategy.
Specifically, parent ideas are selected via tournament selection, while crossover pairing is performed randomly, and the full RBM mechanism is preserved.
Compared to the full model, diversity decreases moderately (from 0.55 to 0.52), while novelty remains unchanged (0.40), and relative quality also decreases from 0.21 to 0.17.
This suggests that RBM plays a dominant role in promoting novelty, while SAC primarily contributes to controlling recombination quality and maintaining balanced semantic coverage during crossover.
The drop in relative quality further indicates that the current SAC design is not merely diversity-oriented, but also supports the generation of more competitive offspring under the automatic comparison protocol.
At the same time, the remaining gap suggests that semantic-aware pairing is still heuristic, and that more adaptive or quality-aware parent matching strategies remain an important direction for future work.

These ablation results help clarify the functional roles of RBM and SAC within the current automatic evaluation setup, but they should not be interpreted as definitive validation of real-world scientific usefulness.

\begin{figure*}[t]
  \centering
  \includegraphics[width=0.9\textwidth]{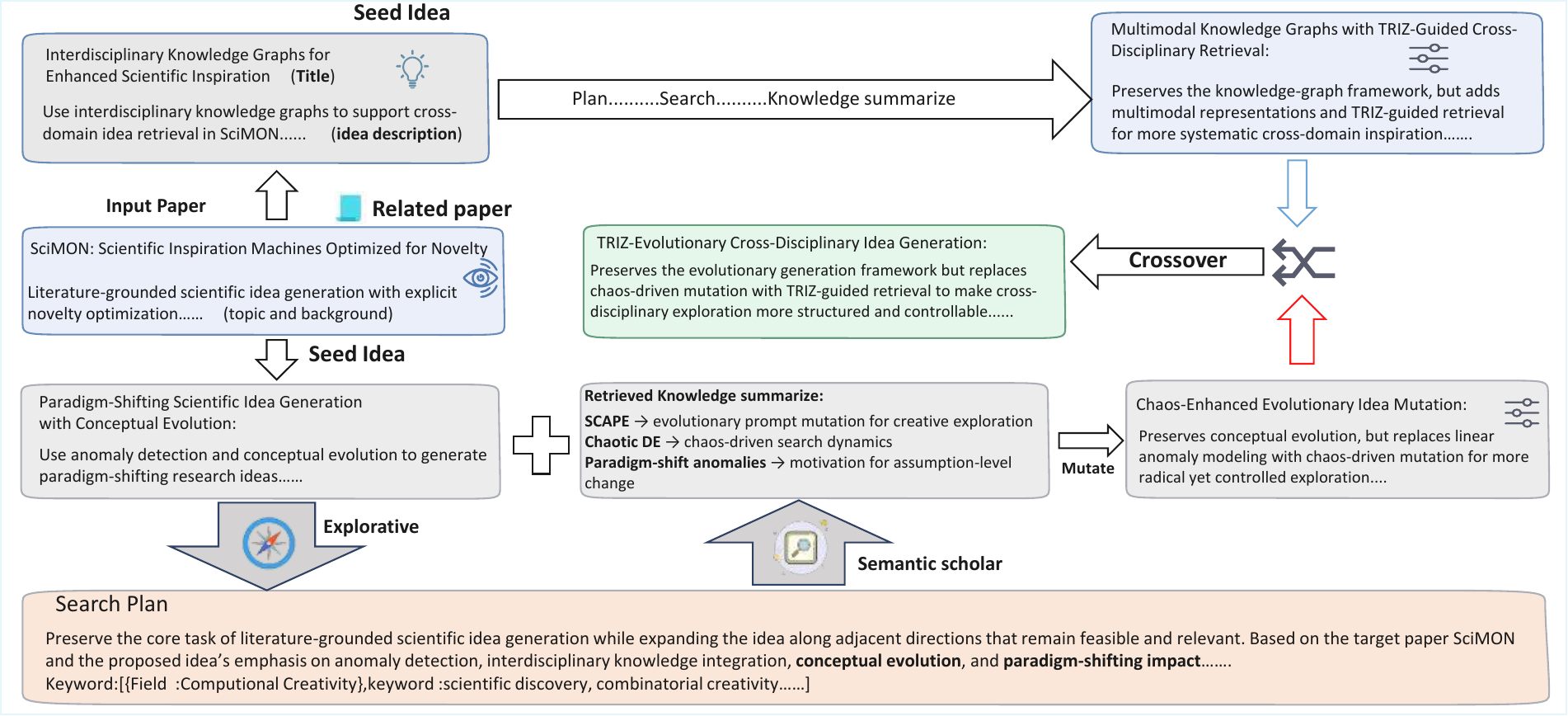}
  \caption{A case study of evolutionary candidate idea development in \textsc{EvoGens}. Starting from a common input paper, two seed ideas are evolved through retrieval-grounded mutation and crossover. For mutation, the figure explicitly shows the generated search plan, the retrieved knowledge, and the resulting mutated idea, illustrating that idea transformation is guided by strategy-specific planning rather than random perturbation. The final crossover offspring preserves complementary components from both parents, demonstrating structured conceptual recombination for early-stage research exploration.}
  \label{fig:case-study}
\end{figure*}

\subsection{Case Study}

Figure~\ref{fig:case-study} presents a representative evolutionary case in \textsc{EvoGens}, showing two seed ideas, the retrieval-grounded mutation process, and the final crossover offspring.
Unlike random perturbation, mutation is explicitly mediated by a generated search plan and the retrieved external knowledge.

In the illustrated mutation branch, the system first assigns a mutation strategy and generates a search plan specifying retrieval fields and keywords.
This plan is executed through Semantic Scholar, and the retrieved literature is summarized as compressed knowledge cues beside the mutated idea.
Guided by this process, the mutated idea \emph{``Chaos-Enhanced Evolutionary Idea Mutation''} preserves the conceptual evolution framework while replacing linear anomaly modeling with chaos-driven mutation, enabling more radical yet still controlled exploration.

The final offspring, \emph{``TRIZ-Evolutionary Cross-Disciplinary Idea Generation''}, is produced by recombining complementary components from two parent ideas.
It preserves the evolutionary generation framework from one branch while incorporating TRIZ-guided cross-disciplinary retrieval from the other, making cross-disciplinary exploration more structured and controllable.
This case shows that mutation in \textsc{EvoGens} is retrieval-grounded instead of random, and that crossover performs mechanism-level conceptual recombination rather than surface-level text fusion.

\section{LIMITATIONS}
Despite its effectiveness, \textsc{EvoGens} has several limitations that warrant further investigation.
(1)~\textit{Heuristic Scoring Design.}~
Although multi-dimensional criteria and aggregation strategies are employed to improve stability, such scores serve only as heuristic fitness proxies for within-population control rather than as optimal or theoretically grounded fitness functions from an evolutionary computation perspective.
Designing more robust, interpretable, and scientifically aligned evaluation functions remains an important challenge.
We also note that this version does not include a dedicated analysis of scorer variance across repeated LLM evaluations or multi-run search stability. Since both generation and evaluation rely on LLM-based components, assessing such variability remains an important direction for future work.
(2)~\textit{Evaluation Scope.}~
Current evaluation of \textsc{EvoGens} is limited to scientific idea generation tasks in the \textit{natural language processing (NLP)} domain.
Although this setting is sufficient to show the effectiveness of the proposed framework, further validation in additional research domains remains an important direction for future work.
More broadly, all automatic evaluations used in this work provide proxy evidence rather than definitive validation of scientific usefulness.
(3)~\textit{Lack of Human Expert Evaluation.}~
Due to practical constraints, we did not involve human evaluation of generated ideas.
As in existing work, the current assessment mainly relies on automated quality metrics and comparative ranking signals, which may not fully capture the true scientific value, feasibility, or long-term usefulness of research ideas.
Incorporating expert judgment remains necessary to better validate real-world scientific utility.

In addition, we do not provide a detailed analysis of computational cost or initialization sensitivity in this version. Although the current settings are sufficient for controlled comparison, understanding these trade-offs more systematically would further strengthen the framework.

\section{ETHICS STATEMENT}

This work proposes \textsc{EvoGens}, an LLM-based evolutionary framework for scientific idea generation.
The study does not involve human participants, personal data, or sensitive information, and therefore does not raise direct concerns regarding privacy or data protection.
A potential ethical consideration concerns the misuse of automatically generated research ideas.
Generated ideas may be incomplete, incorrect, or infeasible, and should not be regarded as validated scientific contributions without expert review.
Accordingly, \textsc{EvoGens} is designed as an assistant tool to support human researchers rather than an autonomous decision-making system. The framework grounds idea generation in retrieved academic literature and employs multi-dimensional quality assessment to reduce unverifiable or hallucinated content.
Overall, \textsc{EvoGens} aims to augment, rather than replace, human scientific creativity.
When used responsibly, it can support early-stage scientific exploration without introducing additional ethical risks.



\bibliographystyle{model1-num-names}

\bibliography{reference}

\appendix
\section{Additional Analysis of Evolution Dynamics}

\label{appendix:Turn-analysis}

\begin{figure}
  \centering
  \small
  \includegraphics[width=\columnwidth]{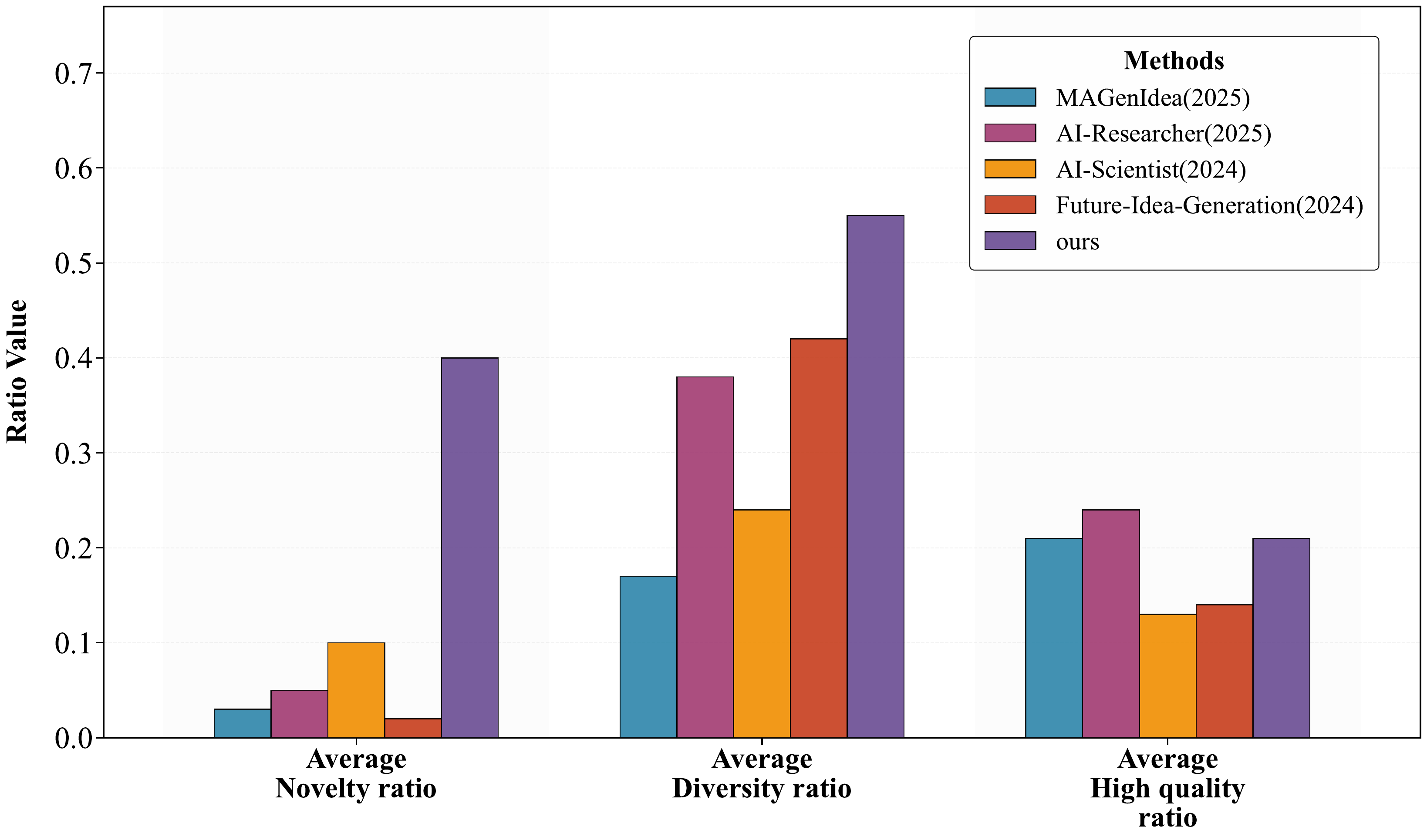}
  \caption{Comparison with baseline methods. \textsc{EvoGens} consistently outperforms baselines in both novelty and diversity.}
  \label{fig:comparison}
\end{figure}

Figure~\ref{fig:comparison} presents a bar-chart comparison between \textsc{EvoGens} and baseline methods in terms of novelty, diversity, and quality.
Overall, \textsc{EvoGens} demonstrates clear advantages in both novelty and diversity,
indicating a stronger capability to explore and maintain a diverse idea space.
Although its quality score is not the highest among all methods,
it remains within a competitive and acceptable range,
suggesting that the gains in exploration do not come at the cost of severely degraded idea quality.

\begin{figure}
  \centering
  \includegraphics[width=\columnwidth]{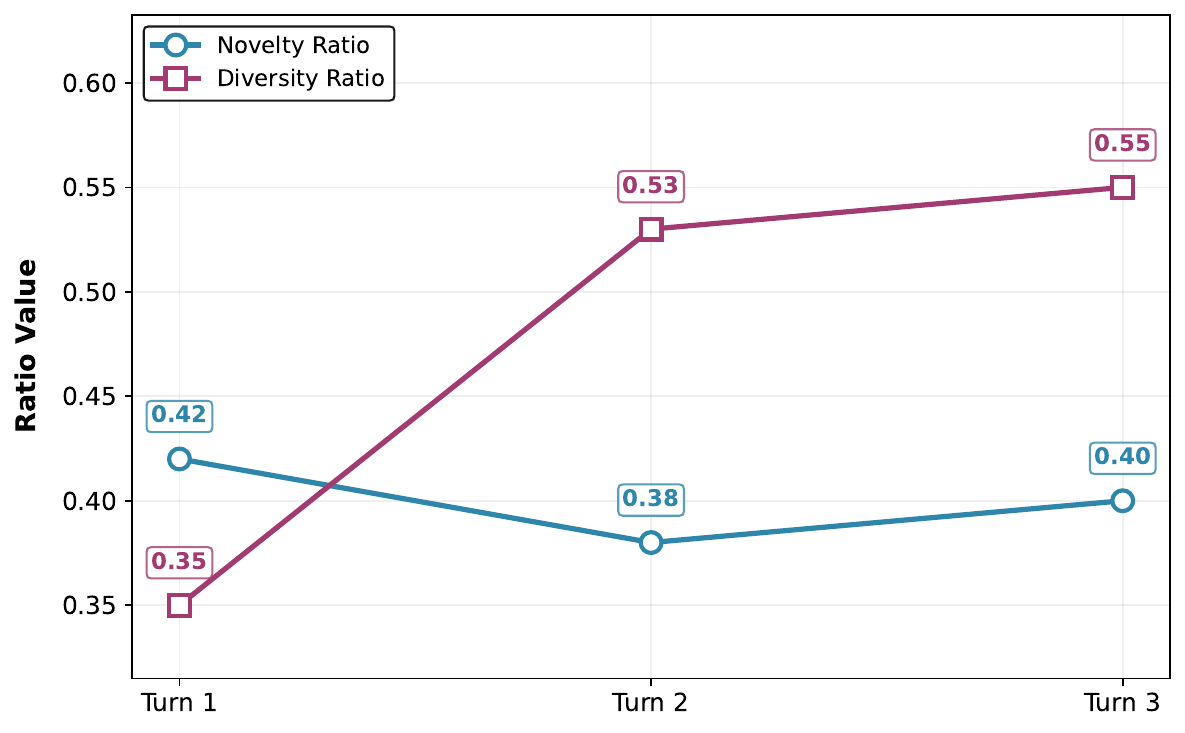}
  \caption{Novelty and Diversity vary with iterations}
  \label{fig:turn}
\end{figure}

In addition, we analyze how novelty and diversity evolve across generations for \textsc{EvoGens}
(see Figure~\ref{fig:turn}). The results show that novelty stabilizes after the first evolutionary turn,
while diversity continues to increase across subsequent turns,
with diminishing gains in later iterations.
This trend suggests that early evolution primarily contributes to expanding the idea space,whereas later turns help consolidate and redistribute ideas
to achieve a more diverse yet stable population.
\subsection{Turn-level Evolution Analysis}
We analyze how novelty and diversity evolve across generations for \textsc{EvoGens} and its ablated variants.
Table~\ref{tab:turn_analysis} reports the novelty and diversity after each evolutionary turn.

\subsection{Effect of Parent Pool Size}
\begin{table}[htbp]
\centering
\renewcommand{\arraystretch}{1.15}
\caption{Effect of parent pool size on novelty and diversity.
The parent pool size controls the number of candidate parents used for crossover in each generation.}
\label{tab:parent_size}

\begin{tabular*}{\columnwidth}{@{\extracolsep{\fill}}lcc@{}}
\hline
Parent Pool Size & Novelty $\uparrow$ & Diversity $\uparrow$ \\
\hline
20 & 0.38 & 0.58 \\
30 & 0.38 & 0.55 \\
40 & 0.31 & 0.54 \\
\hline
\end{tabular*}
\end{table}

We investigate the effect of the parent pool size used during crossover.
As shown in Table~\ref{tab:parent_size}, using an excessively large parent pool leads to a noticeable degradation in novelty, while smaller to moderate pool sizes maintain stable novelty and comparable diversity.
This suggests that overly broad parent selection may introduce redundant or weakly related recombinations, reducing the effectiveness of crossover in generating novel ideas.
Based on this observation, we adopt a moderate parent pool size in all experiments to ensure stable evolutionary behavior.

\section{Prompts}
\label{sec:appendix_PROMPT}
This section mainly introduces the prompts used by \textsc{EvoGens}
in the generation and evolution of ideas.
Specifically, Tables~\ref{tab:conservative_mutation_prompt},
\ref{tab:exploratory_prompt}, and~\ref{tab:radical_prompt}
present the prompts used for the three mutation plans,
corresponding to conservative, exploratory, and radical evolution.
Table~\ref{tab:idea_eval_prompt} details the prompt for idea evaluation,
while Table~\ref{tab:idea_mutate_prompt} summarizes the prompt
used to guide mutation generation.
Finally, Table~\ref{tab:crossover_prompt} provides the prompt
employed in the semantic-aware crossover stage.

\begin{table*}[t]
\centering
\small
\setlength{\tabcolsep}{6pt}
\caption{Prompt used for conservative mutation retrieval planning.}
\begin{tabular}{|p{16cm}|}
\hline
\textbf{You are a research refinement specialist for conservative mutation in evolutionary idea generation.}
Your role is to identify literature that can strengthen, validate, and locally refine a high-potential research idea without changing its core conceptual identity.
Focus on evidence-grounded improvement rather than conceptual expansion.
Avoid recommendations that substantially alter the main research assumption, problem framing, or innovation direction. \\[6pt]

Given a high-ranked research idea, develop a retrieval plan for \textbf{Conservative Mutation}. \\[6pt]

\textbf{Context:} \\
- Research Idea: \{\} \\
- Target Paper: \{\} \\
- Target Paper Abstract: \{\} \\[6pt]

\textbf{Conservative Mutation Objective} \\
This idea is already promising and should be refined rather than redirected.
Your goal is to identify literature that can support \textbf{local, evidence-grounded improvement} while preserving the core research concept. \\[6pt]

\textbf{Retrieval Goals} \\
Search for literature that can help: \\
1. \textbf{Strengthen the existing methodology} with established techniques \\
2. \textbf{Validate the current direction} through closely related successful studies \\
3. \textbf{Improve implementation details, robustness, or evaluation design} \\
4. \textbf{Introduce complementary methods} that enhance the idea without changing its main assumption or scope \\[6pt]

\textbf{Allowed Adjustments} \\
- methodological refinement \\
- improved evaluation protocols \\
- implementation guidance \\
- local scope clarification \\
- robustness enhancement \\[6pt]

\textbf{Avoid} \\
Do NOT search for literature that mainly supports: \\
- paradigm-shifting changes to the core idea \\
- replacement of the central research assumption \\
- large scope transfer to distant domains \\
- highly speculative or weakly grounded alternatives \\
- generic inspiration with unclear relevance \\[6pt]

\textbf{Retrieval Preference} \\
Prioritize: \\
- closely related literature \\
- validated methods \\
- successful implementations \\
- benchmark and evaluation practices \\
- practically useful refinements \\
\hline
\end{tabular}

\label{tab:conservative_mutation_prompt}
\end{table*}

\begin{table*}[t]
\centering
\small
\setlength{\tabcolsep}{6pt}
\caption{Prompt used for explorative mutation retrieval planning.}
\begin{tabular}{|p{16cm}|}
\hline
\textbf{You are a research exploration specialist for explorative mutation in evolutionary idea generation.}
Your role is to identify literature that enables controlled conceptual expansion of a promising research idea.
You should explore adjacent directions, alternative methods, and moderate reframings while preserving feasibility and partial continuity with the original research concept.
Focus on balanced exploration rather than radical departure.
Avoid suggestions that completely replace the original problem identity or lead to highly speculative, weakly grounded directions. \\[6pt]

Given a mid-ranked research idea, develop a retrieval plan for \textbf{Explorative Mutation}. \\[6pt]

\textbf{Context:} \\
- Research Idea: \{\} \\
- Target Paper: \{\} \\
- Target Paper Abstract: \{\} \\[6pt]

\textbf{Explorative Mutation Objective} \\
This idea has potential but still requires broader conceptual expansion.
Your goal is to identify literature that supports \textbf{balanced exploration}: expanding the idea into adjacent directions while preserving feasibility and partial continuity with the original concept. \\[6pt]

\textbf{Retrieval Goals} \\
Search for literature that can help: \\
1. \textbf{Diversify the methodology} through alternative but plausible technical approaches \\
2. \textbf{Extend the idea} into adjacent or emerging application domains \\
3. \textbf{Reframe the problem} from a different but still related perspective \\
4. \textbf{Relax existing constraints} in a controlled way to broaden the search space \\
5. \textbf{Introduce complementary perspectives} that may improve both novelty and practical value \\[6pt]

\textbf{Allowed Shifts} \\
- alternative methods within a related paradigm \\
- moderate scope expansion \\
- adjacent-domain transfer \\
- problem reformulation with preserved relevance \\
- controlled extension of assumptions or constraints \\[6pt]

\textbf{Avoid} \\
Do NOT search for literature that mainly supports: \\
- complete replacement of the original problem identity \\
- highly speculative or weakly grounded research directions \\
- distant cross-domain jumps with little conceptual continuity \\
- disruptive paradigm changes that are more appropriate for radical mutation \\
- generic inspiration without a clear path back to the original idea \\[6pt]

\textbf{Retrieval Preference} \\
Prioritize: \\
- adjacent research areas \\
- feasible alternative methods \\
- transferable techniques \\
- moderate scope expansion \\
- literature that opens new directions without abandoning the original concept \\
\hline
\end{tabular}

\label{tab:exploratory_prompt}
\end{table*}

\begin{table*}[t]
\centering
\small
\setlength{\tabcolsep}{6pt}
\caption{Prompt used for radical mutation retrieval planning.}
\begin{tabular}{|p{16cm}|}
\hline
\textbf{You are a research reframing specialist for radical mutation in evolutionary idea generation.}
Your role is to identify literature that can help substantially reframe a weak or stagnating research idea by challenging its core assumptions and introducing distant but conceptually transferable perspectives.
Focus on controlled long-range exploration rather than arbitrary topic drift.
Seek non-obvious alternatives that can break local convergence, but ensure that the retrieved directions still admit a plausible path of transfer back to the original research problem. \\[6pt]

Given a low-ranked research idea, develop a retrieval plan for \textbf{Radical Mutation}. \\[6pt]

\textbf{Context:} \\
- Research Idea: \{\} \\
- Target Paper: \{\} \\
- Target Paper Abstract: \{\} \\[6pt]

\textbf{Radical Mutation Objective} \\
This idea may benefit from substantial reframing in order to escape local convergence.
Your goal is to identify literature that enables \textbf{controlled long-range exploration} by challenging core assumptions, introducing distant alternative paradigms, or transferring mechanisms from semantically distant but conceptually relevant domains. \\[6pt]

\textbf{Retrieval Goals} \\
Search for literature that can help: \\
1. \textbf{Challenge the core assumptions} of the current idea \\
2. \textbf{Introduce distant but transferable paradigms} that solve related problems differently \\
3. \textbf{Reveal non-obvious mechanisms} from outside the immediate research neighborhood \\
4. \textbf{Reframe the problem} under alternative objectives, constraints, or value functions \\
5. \textbf{Open a path toward disruptive but interpretable reformulation} \\[6pt]

\textbf{Allowed Shifts} \\
- substantial assumption change \\
- distant-domain method transfer \\
- paradigm substitution \\
- unconventional problem framing \\
- alternative evaluation or optimization perspectives \\[6pt]

\textbf{Avoid} \\
Do NOT search for literature that mainly supports: \\
- purely incremental refinements \\
- obvious extensions within the same local neighborhood \\
- distant analogies with no plausible transfer path \\
- metaphorical or decorative cross-domain links without methodological relevance \\
- arbitrary novelty that cannot be reconnected to the original problem \\[6pt]

\textbf{Retrieval Preference} \\
Prioritize: \\
- semantically distant but conceptually transferable domains \\
- competing paradigms \\
- non-obvious mechanisms with interpretable reuse potential \\
- literature that can help break local convergence without collapsing coherence \\
\hline
\end{tabular}

\label{tab:radical_prompt}
\end{table*}

\begin{table*}[t]
\centering
\footnotesize
\setlength{\tabcolsep}{4pt}
\renewcommand{\arraystretch}{1.02}
\caption{Prompt used for literature-grounded idea mutation.}
\begin{tabular}{|p{0.92\textwidth}|}
\hline
\textbf{You are an expert AI researcher who performs literature-grounded idea mutation.}
Your task is to refine or transform an existing research idea by incorporating newly retrieved knowledge, while preserving a clear conceptual connection to the old idea and the target paper. \\[4pt]

\textbf{Mutation Procedure} \\
Generate new candidate ideas by following these steps: \\
1. Understand the target paper and the old idea, and identify which core components of the old idea should be preserved. \\
2. Understand the newly retrieved knowledge, and identify which insights, methods, assumptions, or evaluation strategies can be transferred. \\
3. Propose improved ideas that remain grounded in the target paper and traceably evolve from the old idea, rather than replacing it with an unrelated proposal. \\[4pt]

Please first generate \textbf{5 possible mutated ideas}, analyze the strengths and weaknesses of each, and then select the \textbf{best 2 final ideas}. \\
In the thinking step, explicitly explain: \\
- what is preserved from the old idea, \\
- what is changed using the new knowledge, \\
- why the resulting idea is a meaningful mutation rather than a completely unrelated proposal. \\[4pt]

\textbf{Input Data Description} \\
It is important to understand the target paper, the old idea, and the new knowledge: \\
- The target paper defines the main research topic and scope. \\
- The old idea is the current idea to be improved or transformed. \\
- The new knowledge consists of retrieved literature that provides inspiration for mutation. \\[4pt]

\textbf{Mutation Objective} \\
Your goal is to produce \textbf{literature-grounded mutated ideas}. The final ideas should: \\
1. remain aligned with the target paper's research topic, \\
2. preserve a traceable connection to the old idea, \\
3. incorporate useful insights from the new knowledge, \\
4. introduce non-trivial but coherent changes in motivation, method, scope, or rationale. \\[4pt]

\textbf{Format} \\
In \texttt{<JSON>}, provide the final idea list in JSON format, where every idea contains: \\
- \texttt{"Title"}: concise title \\
- \texttt{"thinking"}: explain what is preserved from the old idea, what is changed, which retrieved knowledge inspired the mutation, and why the result is coherent and valuable \\
- \texttt{"idea"}: the complete mutated research idea \\
- \texttt{"motivation"}: why this mutation is necessary and what new gap or opportunity it addresses \\
- \texttt{"method"}: how the mutated idea would be implemented and evaluated \\
- \texttt{"rationale"}: why this mutation makes sense theoretically and practically \\[4pt]

\textbf{Requirements} \\
1. The new idea should stay within the target paper's topic and improve or extend the old idea. \\
2. The new idea must be meaningfully different from the old idea, while preserving conceptual continuity. \\
3. The retrieved knowledge is used as inspiration for mutation, not as text to be copied or directly cited. \\
4. Avoid generating a completely unrelated new proposal. \\
5. Please think step by step. \\
6. Only the final 2 ideas should be returned in JSON format. \\[4pt]

\textbf{Input:} \\
- target paper title: \texttt{\{t\_title\}} \\
- target paper abstract: \texttt{\{t\_abstract\}} \\
- old\_idea: \texttt{\{seed\_idea\}} \\
- new\_knowledge: \texttt{\{new\_knowledge\}} \\[4pt]

\textbf{Output:} \\
\texttt{\#\#\# Thinking} \\
\texttt{<Explain the target paper topic, what is preserved from the old idea, what is changed using the new knowledge, and why the mutation is coherent and valuable>} \\[4pt]

\texttt{\#\#\# 5 possible mutated ideas (strength and weakness analysis)} \\
\texttt{<MarkDownFormat For 5 Possible Results>} \\[4pt]

\texttt{\#\#\# Final 2 Mutated Ideas} \\
\texttt{<JSON>} \\
\hline
\end{tabular}
\label{tab:idea_mutate_prompt}
\end{table*}

\begin{table*}[t]
\centering
\small
\setlength{\tabcolsep}{6pt}
\caption{Prompt used for mechanism-level crossover in idea generation.}
\begin{tabular}{|p{16cm}|}
\hline
\textbf{You are an expert research innovator specializing in mechanism-level idea crossover.}
Your task is to generate new research concepts by recombining the structured conceptual components of two parent ideas, rather than by blending topics or rewriting phrases.
Focus on conceptual recombination and causal reorganization, not surface-level language fusion. \\[6pt]

\textbf{Crossover Objective} \\
Construct genuinely new research concepts by explicitly recombining structured components from two parent ideas.
The offspring must preserve at least one meaningful component from each parent and introduce a non-trivial change in assumption, method, scope, or contribution. \\[6pt]

\textbf{Procedure} \\
1. Decompose each parent into: core assumption, research motivation, method/mechanism, application scope, and expected contribution. \\
2. Analyze compatibility, conflict, complementarity, and whether direct fusion is feasible. \\
3. Generate offspring only through component-level recombination such as assumption replacement, mechanism transplant, scope transfer, causal workflow recomposition, or contribution reframing. \\
4. If the parents are too distant, keep Parent1 as the structural anchor and import exactly one mechanism-level component from Parent2. \\
5. Generate 5 variants if direct crossover is feasible, or 3 constrained variants if fallback is triggered. \\
6. Select the best 2 variants with distinct innovation dimensions and coherent parent contribution. \\[6pt]

\textbf{Avoid} \\
- simple concatenation of parent ideas \\
- side-by-side restatement of both parents \\
- minor wording variations of the parents \\
- vague fusion language without explicit structural change \\
- superficial hybrids with no coherent transfer path \\[6pt]

\textbf{Evaluation Criteria} \\
- Innovation \\
- Coherence \\
- Groundedness \\
- Impact \\
- Feasibility \\
- Recombination Depth \\[6pt]

\textbf{Input} \\
- Parent1: \{\} \\
- Parent2: \{\} \\[6pt]

\textbf{Output} \\
Return only the final two ideas in JSON format. Each idea should include: \\
- \texttt{"thinking"} \\
- \texttt{"recombination\_type"} \\
- \texttt{"parent1\_contribution"} \\
- \texttt{"parent2\_contribution"} \\
- \texttt{"Title"} \\
- \texttt{"idea"} \\
- \texttt{"motivation"} \\
- \texttt{"method"} \\
- \texttt{"rationale"} \\
\hline
\end{tabular}

\label{tab:crossover_prompt}
\end{table*}

\begin{table*}[t]
\centering
\small
\setlength{\tabcolsep}{6pt}
\caption{Scientific discovery theories used in the idea generation prompt.}
\begin{tabular}{|p{16cm}|}
\hline
\textbf{General laws and methodologies of scientific discovery from the perspective of the philosophy of science} \\[6pt]

1. \textbf{Define New Scientific Problems}. 
Theoretical Basis: Kuhn’s paradigm theory, Laudan’s problem-solving model, Nichols’s problem-generation theory. 
Method: Identify anomalies in existing theories; explore theoretical boundaries and scope of application; integrate interdisciplinary knowledge and discover new problems; re-examine neglected historical problems. \\[6pt]

2. \textbf{Propose New Hypotheses}. 
Theoretical Basis: Pierce’s hypothetical deduction method, Weber’s theory of accidental discovery, Simon’s scientific discovery as problem solving. 
Method: Analogical reasoning; thought experiment; intuition and creative leaps; reductio ad absurdum thinking. \\[6pt]

3. \textbf{Exploring the Limitations and Shortcomings of Current Methods}. 
Theoretical Basis: Popper’s falsificationism, Lakatos’s research program methodology, Feyerabend’s methodological anarchism. 
Method: Critically analyze existing methods; find deviations between theoretical predictions and experimental results; explore the performance of methods under extreme conditions; interdisciplinary comparative methodology. \\[6pt]

4. \textbf{Design and Improve Existing Methods}. 
Theoretical Basis: Laudan’s methodological improvement model, Ziemann’s creative extension theory, Hacking’s experimental system theory. 
Method: Integrate new technologies and tools; improve experimental design and control; improve measurement accuracy and resolution; develop new data analysis methods. \\[6pt]

5. \textbf{Abstract and Summarize the General Laws Behind Multiple Related Studies}. 
Theoretical Basis: Whewell’s conceptual synthesis theory, Carnap’s inductive logic, Glaser and Strauss’s grounded theory. 
Method: Comparative analysis of multiple case studies; identify common patterns and structures; construct conceptual frameworks and theoretical models; formal and mathematical descriptions. \\[6pt]

6. \textbf{Construct and Modify Theoretical Models}. 
Theoretical Basis: Quine’s holism, Lakoff’s conceptual metaphor theory, Kitcher’s unified theory of science. 
Method: Form a balance between reductionism and emergence; develop an interdisciplinary theoretical framework; mathematical modeling and computer simulation; theoretical simplification and unification. \\[6pt]

7. \textbf{Designing Critical Experiments}. 
Theoretical Basis: Duhem--Quine thesis, Bayesian experimental design theory, Mayo’s experimental reasoning theory. 
Method: Designing experiments that can distinguish competing theories; exploring extreme conditions and boundary cases; developing new observation and measurement techniques; designing natural experiments and quasi-experiments. \\[6pt]

8. \textbf{Explaining and Integrating Anomalous Findings}. 
Theoretical Basis: Hansen’s theory of anomalous findings, Sutton’s model of scientific serendipity, Kuhn’s theory of crises and revolutions. 
Method: Revisiting basic assumptions; developing auxiliary hypotheses; exploring new explanatory frameworks; integrating multidisciplinary perspectives. \\[6pt]

9. \textbf{Evaluating and Selecting Competing Theories}. 
Theoretical Basis: Reichenbach’s confirmation theory, Sober’s theory selection criteria, Laudan’s problem-solving progress assessment. 
Method: Comparing theories for explanatory power and predictive power; evaluating the simplicity and elegance of theories; considering the heuristics and research agenda of theories; weighing the empirical adequacy and conceptual coherence of theories. \\[6pt]

10. \textbf{Scientific Paradigm Shift}. 
Theoretical Basis: Kuhn’s theory of scientific revolutions, Toulmin’s model of conceptual evolution, Hall’s dynamic system theory. 
Method: Identify accumulated anomalies and crises; develop new conceptual frameworks; reinterpret and organize known facts; establish new research traditions and practices. \\
\hline
\end{tabular}

\label{tab:scientific_discovery_theory}
\end{table*}

\begin{table*}[t]
\centering
\small
\setlength{\tabcolsep}{6pt}
\caption{Prompt used for scientific idea generation, specifying the task requirements and the expected output format.}
\begin{tabular}{|p{16cm}|}
\hline
\textbf{Follow the steps below to generate a research idea:} \\[6pt]

1. Understanding of the target paper and related papers is essential: \\[3pt]
\hspace*{1em}-- The target paper is the primary research study you aim to enhance or build upon through future research, serving as the central source and focus for identifying and developing the specific research idea. \\[3pt]
\hspace*{1em}-- The referenced papers are studies that the target paper has cited, indicating their direct relevance and connection to the primary research topic you are focusing on, and providing additional context and ideas that are essential for understanding and expanding upon the target paper. \\[6pt]

2. Understanding of the scientific discovery theories is essential. You need to select appropriate theories and combine the information provided by the current paper to come up with creative, influential, and feasible ideas. \\[6pt]

3. Here are 10 general laws and methodologies of scientific discovery from the perspective of the philosophy of science. \\[3pt]

\textit{[The scientific discovery theories shown in Table~\ref{tab:scientific_discovery_theory} are inserted here in the actual prompt.]} \\[6pt]

You can choose one or more of these methodologies and propose a new scientific research idea for the target paper. \\[6pt]

\textbf{Requirements} \\[3pt]
1. Output an idea worth exploring. \\[3pt]
2. You should aim for new research ideas that can potentially win best paper awards at top conferences like ACL, NeurIPS, ICLR, and CVPR. \\[3pt]
3. Skip the research theories that may not well match the target paper; the theory and method you use should make sense and be reasonable for the target paper. \\[3pt]
4. Thinking is your thinking process. Please explain which theory you used in your thinking process. \\[3pt]
5. Please output your thought process. \\[3pt]
6. Please think step by step. \\[6pt]

\textbf{Input:} \\[3pt]
target\_paper: \{\} \\[3pt]
referenced\_papers: \{\} \\[6pt]

\textbf{Please respond in the following format:} \\[3pt]
Thought: \textless THOUGHT\textgreater \\[3pt]
IDEA: \texttt{```json<JSON>```} \\[6pt]

In \textless THOUGHT\textgreater, output your thinking process here, explain why you chose these theories to discover new ideas and why it should have a chance to win the best paper awards at top conferences. \\[3pt]
In \textless JSON\textgreater, provide the new idea with the following fields and provide as many details as possible: \\[3pt]
-- \texttt{"Title"}: A title for the idea, which will be used for the paper writing. \\[3pt]
-- \texttt{"Idea"}: A detailed description of the idea, outlining its significance and potential impact. \\[3pt]
-- \texttt{"Thinking"}: A detailed description of the thinking process, indicating from which scientific discovery theory. \\[3pt]
-- \texttt{"Rationale"}: A detailed description of the rationale for the idea. \\[3pt]
This JSON will be automatically parsed, so ensure the format is precise. \\
\hline
\end{tabular}

\label{tab:scientific_discovery_prompt}
\end{table*}

\begin{table*}[t]
\centering
\small
\setlength{\tabcolsep}{6pt}
\caption{Prompt used for AI-driven research idea evaluation, specifying the criteria and scoring scheme for assessing generated ideas.}
\begin{tabular}{|p{16cm}|}
\hline
\textbf{You are an expert research evaluator specializing in AI-driven scientific idea generation.} \\[6pt]

You are given a candidate research idea that includes a \emph{thinking}, \emph{idea}, \emph{keywords}, and \emph{rationale} section. \\[6pt]

\textbf{Idea to be evaluated:} \{\} \\[6pt]

\textbf{Evaluation Criteria} \\
Evaluate the idea according to the following six dimensions: \\
1. \textbf{Innovativeness} -- originality and creativity relative to existing research trends \\
2. \textbf{Clarity} -- clarity and coherence of the idea presentation \\
3. \textbf{Feasibility} -- practicality and implementability under current research constraints (data, computation, methodology) \\
4. \textbf{Significance} -- potential impact if the idea is successfully developed \\
5. \textbf{Effectiveness} -- logical consistency among the thinking, idea, and rationale components \\
6. \textbf{Generalizability} -- potential for extension, reuse, or serving as a foundation for future research \\[6pt]

\textbf{Scoring Scheme} \\
For each criterion, assign a score from 1 to 10, where: \\
- 1 indicates very poor performance, \\
- 5 indicates moderate quality, and \\
- 10 indicates excellent quality. \\[6pt]

\textbf{Output Requirement} \\
Provide a brief textual summary (2--3 sentences) highlighting the idea’s strengths, weaknesses, and overall assessment. \\
\hline
\end{tabular}

\label{tab:idea_eval_prompt}
\end{table*}

\begin{table*}[t]
\centering
\small
\setlength{\tabcolsep}{6pt}
\caption{Prompt used for unified search-plan generation in the w/o RBM setting, specifying how retrieval fields and search keywords are produced from a given research idea.}
\begin{tabular}{|p{16cm}|}
\hline
\textbf{According to a given research idea, analyze which fields of papers should be retrieved in order to collect comprehensive information and new knowledge to support further research and discovery of new ideas. Please provide the thinking process and search keywords.} \\[6pt]

\textbf{Research idea:} \{\} \\[3pt]
\textbf{Target paper title:} \{\} \\[6pt]

\textbf{Requirements} \\[3pt]
1. The search plan should be developed around the given research idea. \\[3pt]
2. The plan should focus on how to optimize the target paper. \\[3pt]
3. Please provide both the thought process and the search keywords. \\[3pt]
4. Please output the thinking process. \\[6pt]

\textbf{Output Format} \\[3pt]
Thought: \textless THOUGHT\textgreater \\[3pt]
Search Plans: \texttt{```json<JSON>```} \\[6pt]

In \textless THOUGHT\textgreater, explain why you develop such a search strategy. \\[3pt]
In \textless JSON\textgreater, provide a search plan with as much detail as possible, including: \\[3pt]
-- \texttt{"Search Plan"}: The search plan for the given research idea. \\[3pt]
-- \texttt{"Search Fields"}: The list of fields that should be retrieved. \\[3pt]
-- \texttt{"Search\_Keywords"}: The list of keywords for each field (maximum 3 keywords per field), formatted as: \\[3pt]
\texttt{'Search\_Keywords':[{"Field":"field\_name","Keywords":["keyword1","keyword2","keyword3"]}]} \\[6pt]

This JSON will be automatically parsed, so ensure that the format is precise and that each field contains no more than 3 keywords. \\[3pt]
No more than 3 fields. \\
\hline
\end{tabular}

\label{tab:search_keyword_prompt}
\end{table*}





\end{document}